\documentstyle[epsfig]{llncs}

\title{Constraint Categorial Grammars}
\author{ Lu\'{\i}s Damas, Nelma Moreira\\ 
 \small{\{luis,nam\}@ncc.up.pt}}
\institute{LIACC, Universidade do Porto \\Rua
    do Campo Alegre 823, 4150 Porto, Portugal }
\date{\ }

  \newcommand{\lapp}{\backslash} \newcommand{\LFD}{$\Lambda_{\cal
    FD}$} \newcommand{\LFDD}{\Lambda_{\cal FD}}

\newcommand{\llambda}{\lambda\!\!\lambda}

 \newcommand{\rw}{\rightarrow}

\newcommand{\lsbra}{[\![} \newcommand{\rsbra}{]\!]}

 \newcommand{\sfd}[1]{\lsbra
  #1\rsbra\rho}

\newcommand{\Equiv}{\leftrightarrow}

\newtheorem{Th}{Theorem}[section]

 \def\And{\:\wedge\:}  \def\CC{{\cal
    C}} \def\Cat{{\cal C}at} \def\MM{{\cal M}} 
 \def\RR{{\cal RT}} 
 \def\RT{${\cal RT}$} \def\CLG{${\cal CLG}\;$}
\def\CCLG{${\cal CCLG}\;$} 
\def\CLPLFD{${\cal CLP(\LFDD)}\;$}

\def\RwM{\ \longrightarrow_{\cal M}\ } \def\RMs{\ 
  \longrightarrow_{\cal M}^\star\ } \def\fst{{fs\rw\tau}}
\def\tlt{{\tau'\rw\tau}}

\def\comment#1{}

\begin{document}
\maketitle
\begin{abstract}
  Although unification can be used to implement a weak form of
  $\beta$-reduction, several linguistic phenomena are better handled by
  using some form of $\lambda$-calculus. In this paper we present a
  higher order feature description calculus based on a typed
  $\lambda$-calculus. We show how the techniques used in \CLG for
  resolving complex feature constraints can be efficiently extended.
  \CCLG is  a simple formalism, based on categorial grammars, designed
  to test the practical feasibility of such a calculus. 

\end{abstract}
\section{Introduction}

Unification based formalisms show a clear inability to deal in a
natural way with phenomena such as the semantics of coordination,
quantification scoping ambiguity or bound anaphora. As a matter of
fact, although unification can be used to implement a weak form of
$\beta$-reduction, it seems that this kind of phenomena is better
handled by using some form of $\lambda$-calculus
\cite{DalrympleSP91,Pereira90}.  One possibility, which is at the
heart of systems like $\lambda$Prolog\cite{NadathurM88}, is
to extend both the notion of term, to include $\lambda$-abstraction
and application, and the definition of unification to deal with
$\lambda$-terms.  For this extension to be technically sound it is
necessary to require $\lambda$-terms to be well typed.  On the other
hand, it turns out that if instead of using terms we use complex
feature descriptions (where conjunction replaces unification), we
still can follow the same plan to produce a higher-order calculus of
feature descriptions.  \CCLG is a simple formalism, based on
categorial grammars, designed to test the practical feasibility of
such an approach.  The main reason for selecting a categorial
framework for this experiment was that, due to the simplicity of the
categorial framework, it allowed us to concentrate on the constraint
calculus itself. Another reason was also the close historical
relationship between categorial grammars and semantic formalisms
incorporating $\lambda$-abstraction. \CCLG extends categorial grammar
by associating not only a category but also a higher-order feature
description with each well-formed part of speech.  The type of these
feature descriptions are determined by the associated category. Note
also that a derivation leading to an unsatisfiable feature description
is legal.  When compared with other formalisms (for instance,
\cite{ZeevatKC87}) one of the main distinguishing features of \CCLG is
the fact that it computes partial descriptions of feature structures
and not the feature structures themselves.

It is important to notice that this calculus is easily modified to
deal with constraints over finite or rational trees, instead of
feature trees.  Also, the advantages of this kind of calculus, namely
its decidability, over the use of general high-order logic programming
systems, for processing semantic representations in NLP
systems should be obvious. The rest of the paper proceeds as follows.
We start by defining a feature description calculus as an hybrid of
$\lambda$-calculus and feature logics and we present its denotational
semantics. In section \ref{cs} we describe  a complete constraint solver for
higher-order feature descriptions. In section \ref{catgram} 
we define constraint categorial grammars  and briefly present its
implementation. Some final remarks are considered in section \ref{fr}.  
\section{Feature Description Calculus}\label{fdc}

The feature description calculus \LFD\ at the heart of our formalism
\comment{CCLG} is inspired both on the $\lambda$-calculus and on
feature logics \cite{Smolka89,SmolkaT92}. For technical reasons,
namely that we want to ensure the existence of normal forms, it is a
typed calculus. Our base types are {\bf bool} for truth values and
{\bf fs} for feature structures.

Our types are described by
$$\tau ::= {\bf bool} \mid {\bf fs} \rightarrow \tau \mid \tau
\rightarrow \tau'$$
\noindent Note that we exclude {\bf fs} as the type of any feature
description.  This reflects our commitment to compute partial
descriptions of feature structures rather than feature structures.

Now assume we are given a set of {\em atoms} $a$, $b$, \ldots, a set
of {\em feature} symbols $f$, $g$, \ldots, a set of {\em feature
  structure variables} $x$, $y$, \ldots, and, for each type $\tau$, a
set of {\em variables of type $\tau$} $x_{\tau}$, $y_{\tau}$, \ldots.
Then the set of {\em feature descriptions} of type $\tau$ is described
by
\[
\begin{array}{lll}
  e_{\tau} &::=& {\bf true} \mid {\bf false } \mid x_{\tau} \mid
  e_{\tau} \wedge e_{\tau} \mid e_{\tau} \vee e_{\tau} \mid \neg
  e_{\tau} \mid e_{fs\rightarrow\tau}x.p \mid
e_{fs\rightarrow\tau}a \mid e_{\tau'\rightarrow\tau}e_{\tau'}\\ 
e_{bool} &::=& t.p\dot=s \mid t=s\\ e_{fs\rightarrow\tau}
&::=& \lambda x.e_{\tau}\\ e_{\tau'\rightarrow\tau} &::=& \lambda
x_{\tau'}.e_{\tau}
\end{array}
\]
\noindent where $s$ and $t$ denote either atoms or feature structure
variables, and $p$ is a, possibly empty, sequence of feature symbols
denoting a path in a feature structure.

Note that the language thus defined includes both feature logics and a
typed $\lambda$-calculus. 

We import from both theories such notions as
substitution, free and bounded  occurrences of variables, $\alpha-$ and
$\beta-$reductions and $\beta\alpha$-normal form. In particular, a {\em closed feature
description\/} is a feature description with no free variables.
 Moreover, feature constraints of
feature logics, widely used in unification grammars, correspond to a subset of {\em feature descriptions}
of type {\bf bool}, without abstractions or applications.

To define a semantics for the calculus of feature descriptions we
adopt the standard model ${\cal RT}$ of rational trees for feature
structures (see \cite{DamasMV92}) and  associate with each type $\tau$ a semantic domain
$D_{\tau}$ as follows
\[
\begin{array}{lll}
  D_{\bf bool} &=& \{0,1\}\\ D_{{\bf fs} \rightarrow \tau} &=& {\cal
    RT} \rightarrow D_{\tau} \\ D_{\tau'\rightarrow \tau} &=&
  D_{\tau'} \rightarrow D_{\tau}
\end{array}
\]
From this point on a semantics for feature descriptions is defined in
the same way as for feature logics and the typed $\lambda$-calculus by
noting that the standard boolean operations can be extended to all the
semantic domains involved in a component wise fashion, e.g.
$$ (\lambda x.e) \vee (\lambda x.e') =_{def} (\lambda x. e \vee e').$$
\comment{Similarly, for each type $\tau$, ${\bf true}$ and ${\bf false}$ denote
the obvious elements of ${\cal D}_{\tau}$.}  
More precisely, let an
{\em assignment} $\rho$ be a mapping defined on variables, such that
$\rho(x) \in $ \RT\ and $\rho(x_\tau) \in D_\tau$, for each type
$\tau$. As usual, $\rho[d/\sigma]$ denotes the assignment obtained
from $\rho$ by mapping $\sigma$ to $d$.  Let $f^{\cal RT}$, $p^{\cal
  RT}$ and $a^{\cal RT}$ denote the interpretation of features, paths
and atoms in \RT, respectively. Furthermore, let $t^{\cal RT}\rho$ be
$\rho(t)$ if $t$ is a variable and $t^{\cal RT}$ otherwise.  Then, the
semantics of feature descriptions \LFD\ given an assignment $\rho$ is
defined inductively, as follows:
\[
\begin{array}{ll}
\begin{array}{llll}
  \lsbra x_\tau \rsbra\rho&=& \rho(x_\tau)\\ 
\lsbra t.p\dot=s \rsbra\rho &=& \left\{
\begin{array}{l}
  1 \;\;\;if \;\;p^{\cal RT}(t^{\cal RT}\rho)= s^{\cal RT}\rho \\ 0 \;\;\;
  otherwise
\end{array} \right. \\
\lsbra
  t=s\rsbra \rho &=&
 \left\{
\begin{array}{l}
  1 \;\;\;if \;\;t^{\cal RT}\rho= s^{\cal RT}\rho \\ 0 \;\;\;
  otherwise
\end{array} \right. 
\\\end{array}
&
\begin{array}{llll}
  \lsbra \lambda x.e_\tau \rsbra\rho&=& \llambda v.\lsbra e_\tau
  \rsbra\rho[v/x] & (v\;\in\;{\cal RT})\\ 

  \lsbra \lambda x_\tau.e_\tau' \rsbra\rho&=& \llambda v.\lsbra e_\tau'
  \rsbra\rho[v/x_\tau] & (v\; \in\;{\cal D}_\tau)\\ \lsbra
  e_{fs\rightarrow\tau} x.p\rsbra\rho &=& (\lsbra
  e_{fs\rightarrow\tau} \rsbra \rho) \lsbra x.p\rsbra\rho\\ \lsbra
  e_{fs\rightarrow\tau} a \rsbra \rho &=& (\lsbra e_{fs\rightarrow\tau}\rsbra\rho)a^{\cal RT}\\ \lsbra
  e_{\tau\rightarrow\tau'} e_\tau'\rsbra\rho &=&\ (\lsbra
  e_{\tau\rightarrow\tau'} \rsbra\rho)\lsbra e_\tau' \rsbra\rho
\end{array}
\end{array}
\]
\noindent where $\llambda$ denotes function ``abstraction'' in set
theory and $(x\in D)$ means that $D$ is the domain of $x$. For the conjunction operation, we define:
\[
\begin{array}{llll}
  \lsbra e_{bool} \wedge e'_{bool} \rsbra \rho &=& \left\{
  \begin{array}{ll} 1 & \mbox{ if $\lsbra e_{bool} \rsbra\rho=1$ and
      $\lsbra e'_{bool} \rsbra\rho=1$}\\ 0 & \mbox{ otherwise}
\end{array}\right.\\
\sfd{e_{fs\rightarrow\tau} \wedge e'_{fs\rightarrow\tau}} &=& \llambda
v.d \wedge d'\mbox{ where $\begin{array}[t]{ll}
  \sfd{e_{fs\rightarrow\tau}}=\llambda v.d &(v\;\in\;{\cal RT})\\ 
  \sfd{e'_{fs\rightarrow\tau}}=\llambda v.d' & (v\;\in\;{\cal RT})
  \end{array}$}\\
\sfd{e_{\tau'\rightarrow\tau} \wedge e'_{\tau'\rightarrow\tau}}&=&
\llambda v.d \wedge d'\mbox{ where $\begin{array}[t]{ll}
  \sfd{e_{\tau'\rightarrow\tau}}=\llambda v.d &(v\;\in\;{\cal D}_{\tau)'}\\ 
  \sfd{e'_{\tau'\rightarrow\tau}}=\llambda v.d' & (v\;\in\;{\cal D}_{\tau'})
  \end{array}$}
\end{array}
\]
\noindent and analogously for the other boolean operations. If
$1_{bool}\equiv1$, $1_{fs\rw\tau}\equiv\lambda v.1_\tau$, and
$1_{\tau'\rw\tau}\equiv\lambda v. 1_{\tau}$ then the semantics of {\bf
  true}, for each type $\tau$ is defined by:
\[
\begin{array}{lll}
  \sfd{{\bf true}_{bool}}=1_{bool},\\ \sfd{{\bf
      true}_{fs\rw\tau}}=\lambda v. 1_\tau \;(v\;\in\;{\cal RT})\\ 
  \sfd{{\bf true}_{\tau'\rw\tau}}=\lambda v.  1_\tau\;(v\;\in\;{\cal
    D}_\tau)
\end{array}
\]
\noindent and analogously for {\bf false}.

An important property of the feature description calculus is the
existence of normal form under $\beta$-reduction which is a simple
consequence of well-typedness. Another important property is that for
any closed feature description of type $\tau$ we can decide if it
is equivalent to ${\bf false}$. This last property is essentially an
extension of the satisfiability problem for a complete
axiomatization of feature logics. For this reason we will say that a
feature description of type $\tau$ is satisfiable iff its semantics is
not that of ${\bf false}$.
\subsection{Constraint Solver}\label{cs}
Our implementation of the feature description calculus is based on the
reduction to normal form followed by the techniques used in \CLG
\cite{DamasV92,DamasMV92} for resolving complex feature constraints.
In order to face the NP-hardness of the satisfiability problem, our
approach was based in factoring out, in polynomial time, deterministic
information contained in a complex constraint and simplifying the
remaining formula using that information. The deterministic
information corresponds to a conjunction  of (positive) atomic
constraints in {\em solved form}\footnote{
A conjunction of feature constraints $\MM$ is a {\em solved form} if:
\begin{enumerate}
\item every constraint in $\MM$ is of the form 
$x.f \dot= s$ or  $x=s$
\item if $x=s$ is in $\MM$ then $x$ occurs exactly once in $\MM$
\item if $ x.f\dot=s$ and $x.f\dot=t$ are in $\MM$ then $s=t$
\end{enumerate}
Any conjunction of atomic constraints is satisfiable if and only if it
can be reduced to a {\em solved form}
\cite{Smolka89,Maher88a,DamasMV92}}, which we denote by $\MM$. We say
that ${\cal M}$ is a {\em partial model} of $\cal C$ if and only if
every model of $\cal C$ is a model of ${\cal M}$. When every model of
$\MM$ is a model of $\cal C$, but no proper subset of $\MM$ satisfies
this condition, we will say that ${\cal M}$ is a {\em minimal} model
of $\cal C$. By using disjunctive forms it can be proved that any set
of feature constraints $\cal C$ admits at most a finite number of minimal
models\footnote{Actually it is necessary to extend the notion of models
  to include negative atomic constraints, but that will not be addressed here.}.  In \cite{DamasV92,DamasMV92,Moreira95} a rewrite system was
presented that from a complex feature constraint $\CC_0$ produces a
pair $\langle {\cal M},\CC\rangle$, where $\MM$ is {\em solved form}, $\CC$
is smaller than $C_0$ and such that $\RR\models\CC_0\Equiv
\MM\And\CC$ and any minimal model of $\CC_0$ can be
obtained by conjoining a minimal model of $\CC$ with $\MM$. Moreover the rewriting system  is complete in the sense that ${\cal
    M}\And {\cal C}$ is
satisfiable, unless it produces ${\bf false}$ as the
final model.

\medskip

\noindent We now extend that rewrite system to higher-order feature descriptions.
First we give some more characterizations of feature descriptions.
A {\em basic normal description} of type $\tau$ is described by:
\[
\begin{array}{lll}
       e_{\tau} &::=& {\bf true} \mid {\bf false } \mid x_{\tau} \mid x_{\tau} \wedge e_{\tau}  \mid x_{\tau} \vee e_{\tau} \mid \neg e_{\tau} \mid
        x_{{\em fs}\rightarrow\tau}x.p \mid
        x_{{\em fs}\rightarrow\tau}a \mid
       x_{\tau'\rightarrow\tau}e_{\tau'}\\
       e_{\em bool} &::=&  t.p\dot=s \mid t\dot=s\\
       e_{fs\rightarrow\tau} &::=& \lambda x.e_{\tau}\\
       e_{\tau'\rightarrow\tau} &::=& \lambda x_{\tau'}.e_{\tau}
\end{array}
\]

\noindent
Then, every closed feature description in basic normal form will be of
the form
$\lambda \bar{x}_\sigma.e_{bool}$ where $\bar{x}_\sigma$ denotes a
sequence of bound variables of some types and $e_{bool}$ is not an
abstraction. Omitting the $\lambda$ prefix, given a feature description of type {\bf
  bool}, $e_{bool}$, the solver will produce a partial model $\MM$
and a smaller feature description $e_{bool}'$ or {\bf false}:

$$\begin{array}{lll} \langle{\cal M},e_{bool}\wedge
{\bf false}\rangle \rightarrow \langle\bot,{\bf false}\rangle\\ \langle {\cal
  M},e_{bool} \wedge {\bf true} \rangle \rightarrow \langle {\cal
  M},e_{bool} \rangle\\ \langle{\cal M},e_{bool}\wedge s=t\rangle
\rightarrow \langle{\cal M}\wedge s=t,e_{bool}\rangle \\ \langle{\cal
  M},e_{bool}\wedge t.p\dot=s \rangle \rightarrow \langle{\cal M}
\wedge t.p\dot=s,e_{bool}\rangle \\ 
\langle {\cal M},e_{bool}\rangle \rightarrow \langle {\cal
  M},e^\prime_{bool} \rangle\mbox{ if $e_{bool}\RMs e^\prime_{bool}$}
\end{array}$$
 
\noindent with the convention that after each application of one of
the rewrite rules the new partial model is reduced to solved form (or
{\bf false}). The complete rewrite system $\RwM$ is:

\begin{eqnarray}
\begin{array}{lll}
(\lambda x.e_\tau)x.p &\RwM & e_\tau[x.p/x]\\
(\lambda x.e_\tau)a &\RwM & e_\tau[a/x]\\
(\lambda x_{\tau'}.e_\tau)e'_{\tau'} &\RwM &
e_\tau[e'_{\tau'}/x_{\tau'}]
\end{array} &
\begin{array}{lll}
\neg (\lambda x.e)&\RwM&\lambda x. \neg e\\
(\lambda x.e)\wedge(\lambda x.e')&\RwM&\lambda x.e \wedge e'\\
(\lambda x.e)\vee(\lambda x.e')&\RwM&\lambda x.e \vee e
\end{array}
\end{eqnarray}
\begin{eqnarray}\label{appl}
\begin{array}{llll}
e_\tau \wedge x_\tau&\RwM&x_\tau \wedge e_\tau&\mbox{if $e_\tau$ is
  not a variable}\\
e_\tau \wedge (x_\tau \wedge e_\tau') &\RwM& x_\tau \wedge (e_\tau
\wedge e_\tau')& \mbox{if $e_\tau$ is not a variable}\\
(e_\tau \wedge e_\tau') \wedge e_\tau''&\RwM&e_\tau \wedge (e_\tau'
\wedge e_\tau'')\\
e_\tau \vee x_\tau&\RwM&x_\tau \vee e_\tau&\mbox{if $e_\tau$ is
  not a variable}\\
e_\tau \vee (x_\tau \vee e_\tau') &\RwM& x_\tau \vee (e_\tau
\vee e_\tau')& \mbox{if $e_\tau$ is not a variable}\\
(e_\tau \vee e_\tau') \vee e_\tau''&\RwM&e_\tau \vee (e_\tau'
\vee e_\tau'')\\
(x_{fs\rw\tau} \wedge e_{fs\rw\tau})x.p &\RwM &
 (x_{fs\rw\tau})x.p \wedge (e_{fs\rw\tau})x.p\\
(x_{fs\rw\tau} \wedge e_{fs\rw\tau})a&\RwM &
 (x_{fs\rw\tau})a\wedge (e_{fs\rw\tau})a\\
(x_{\tau'\rw\tau} \wedge e'_{\tau'\rw\tau})e_\tau &\RwM &
 (x_{\tau'\rw\tau})e_\tau \wedge (e'_{\tau'\rw\tau})e_\tau\\
(x_{fs\rw\tau} \vee e_{fs\rw\tau})x.p &\RwM &
 (x_{fs\rw\tau})x.p \vee (e_{fs\rw\tau})x.p\\
(x_{fs\rw\tau} \vee e_{fs\rw\tau})a&\RwM &
 (x_{fs\rw\tau})a\vee (e_{fs\rw\tau})a\\
(x_{\tau'\rw\tau} \vee e'_{\tau'\rw\tau})e_\tau &\RwM &
 (x_{\tau'\rw\tau})e_\tau \vee (e'_{\tau'\rw\tau})e_\tau\\
\end{array}
\end{eqnarray}
\begin{eqnarray}\label{abs2}
\begin{array}{llll}
\lambda x.e_\tau &\RwM & \lambda x.e'_\tau &
        \mbox{   if $e_\tau{\RMs} e'_\tau$}\\
\lambda x_{\tau'}.e_\tau &\RwM & \lambda x_{\tau'}.e'_\tau &
        \mbox{   if $e_\tau\RMs e'_\tau$}\\
(x_{\tau'\rw\tau} e_\tau) &\RwM& (x_{\tau'\rw\tau}e'_\tau) &
        \mbox{   if $e_\tau\RMs e'_\tau$}
\end{array}
\end{eqnarray}
\begin{eqnarray}\label{truefalse}
\begin{array}{ll}
\begin{array}{lll}
false_\fst x.p &\RwM & false_\tau \\
false_\fst a &\RwM & false_\tau \\
false_\tlt e_{\tau'}&\RwM & false_\tau \\
true_\fst x.p &\RwM & true_\tau \\
true_\fst a &\RwM & true_\tau \\
true_\tlt e_{\tau'}&\RwM & true_\tau \\
false_\tau \wedge  e_\tau &\RwM &false_\tau
\end{array}
&
\begin{array}{lll}
true_\tau \wedge  e_\tau &\RwM &e_\tau \\
e_\tau \wedge false_\tau &\RwM &false_\tau \\
e_\tau \wedge  true_\tau &\RwM & e_\tau \\
false_\tau \vee  e_\tau &\RwM & e_\tau \\
true_\tau \wedge  e_\tau &\RwM &true_\tau \\
e_\tau \vee  false_\tau &\RwM &e_\tau \\
e_\tau \wedge  true_\tau &\RwM &true_\tau 
\end{array}
\end{array}
\end{eqnarray}
\begin{eqnarray}
  \label{rwm}
\begin{array}{llll}
e_\tau & \longrightarrow_\bot & false_\tau \\
x & \RwM & t & \mbox{if $x=t\in \MM$}\\
a.p\dot=s &\RwM & false\\
x.p\dot=s &\RwM & t=s &\mbox{if $x.p\dot=t \in \MM$}\\
a=b &\RwM & false \\
t=t &\RwM & true \\
x=t \wedge e_{bool} &\RwM & x=t \wedge e'_{bool}&
    \mbox{if $e_{bool}\ \longrightarrow_{{\cal M}
         \wedge x = t}^\star\ e'_{bool}$}\\
x=t \wedge e_{bool} & \RwM&false &
\mbox{if ${\cal M} \wedge x = t\, \rw\, \bot$}\\
x.p\dot=t \wedge e_{bool} &\RwM & x.p\dot=t \wedge e'_{bool}&
    \mbox{ if $e_{bool}\ \longrightarrow_{{\cal M}
         \wedge x.p\dot= t}^\star\ e'_{bool}$}\\
x.p\dot=t \wedge e_{bool}&\RwM& false & 
\mbox{if ${\cal M} \wedge x.p= t\, \rw\, \bot$}
\end{array}
\end{eqnarray}
\begin{eqnarray}
  \label{dist}
\begin{array}{llll}
(e_\tau \vee e'_\tau) \wedge e''_\tau&\RwM & (e_\tau \wedge e''_\tau)
\vee  (e'_\tau \wedge e''_\tau)&\parbox[t]{4cm}{if both $e_\tau$   and
  $e'_\tau$  are ${\cal M}$-dependent with $e''_\tau$}
\end{array}
\end{eqnarray}
\noindent
We assume that $\alpha$-reductions will be performed whenever
necessary. For
simplicity we omitted the rules concerning negation.
The rewrite system is divided in six groups, each one dealing with:
(1) $\beta$-reduction (where $e[d/x]$ denotes the substitution in $e$ of $x$ for $d$), abstraction and boolean operations for higher
order types; this rules are applied before any other rule (2)
application and boolean operations (3) rewrite
inside abstractions and applications (4) {\bf false}
and {\bf true}; (5) feature description of type {\bf
  bool}, $e_{bool}$; this rules essentially  correspond to the feature
constraint rewrite system in \cite{DamasMV92} (6) distributive law;
this rule  must apply only when both $e_\tau$ and $e'_\tau$ have
variables in common with $e''_\tau$,  eventually through
``bindings'' in $\MM$ \footnote{The notation of $\MM$-dependence coincides with
the one for complex feature constraints \cite{DamasMV92}, if $x\in
c$ means $x$ occurs free in $c$. Given two constraints
$c_1$ and $c_2$ and a model $\MM$, $c_1$ and $c_2$
are {\em $\MM$-dependent} if and only if $Var_{\MM}(c_1) \cap
Var_{\MM}(c_2) \not= \emptyset$, where $Var_{\MM}(c)$ is
the smallest set satisfying:
if $x \in c$, then $x \in Var_{\MM}(c)$;
if $x \in Var_{\MM}(c)$ and $x.f \doteq z \in \MM$,
then $z \in Var_{\MM}(c)$.}. If this last rule is omitted, the rewrite process
becomes polynomial although incomplete.
\begin{Th}
  Given a closed feature description $e_\tau$
  the rewrite  system is correct, terminating and complete
  in the sense that  $e_\tau$ is  satisfiable  unless  {\bf false}  is
  produced. Moreover the final feature description is in basic normal form.
\end{Th}

\noindent For a proof of the  above results see \cite{Moreira95}.
\section{Constraint Categorial Grammar}\label{catgram}

In this section we show how the expressiveness of categorial grammars
can be augmented using feature descriptions.

We will use a basic (rigid) categorial grammar ($CG$), consisting of a set of
categories, a lexicon which assigns categories to words and a calculus
which determines the set of admissible category combinations.
Given a set of basic categories $\Cat_0$ we define recursively the set of
categories $\Cat$ by: the elements of $\Cat_0$ are categories;
if $A$ and $B$ are categories then $A/B$ and $A\lapp B$ are
categories.
Some unary (lexical) rules (lifting, division, etc)  will be 
 added to provide a flexible $CG$  which can cope with discontinuity
and other linguistic phenomena. Semantically these rules allow
functional abstraction over displaced or missing elements.

A Constraint Categorial Grammar is a tuple  $<Cat_0,\Upsilon,Lexicon,Rules>$
where
\begin{enumerate}
 \item $Cat_0$ is a set of base categories
\item  $\Upsilon$ is a map
which associates with each  category $C$ a type $\Upsilon(C)$ and 
satisfies
$$\Upsilon(A/B) = \Upsilon(B\lapp A) = \Upsilon(B) \rightarrow
\Upsilon(A) $$
\item $Lexicon$ is a set of triples
$<w,A,c>$, where $w$ is a word, $A$ a category and  $c$ is a feature
description of type $\Upsilon(A)$
 \item  $Rules$ is the set of
inference rules to combine pairs $A-c$ of syntactic categories and 
feature descriptions (semantic representation).
\end{enumerate}

 The inference rules used in the current grammars are:
\[
\begin{array}{lll}
(app/)&\frac{A/B-c_f\;\;\; B-c_b}{A-(c_fc_b)}& \mbox{if $c_fc_b$ is satisfiable} \\
(app\lapp)&\frac{B-c_b\;\;\;B\lapp A-c_f}{A-(c_fc_b)}
  &\mbox{if $c_fc_b$ is satisfiable}
\end{array}
\]
plus a set of unary rules.

\subsection{A sample grammar}\label{sampgram}
In figure \ref{fig1}. is given a fragment of an English grammar 
written in \CCLG. We use '$\lapp$' for
'$\lambda$', '\&' for '$\wedge$' and '$\mid$' for '$\vee$'. All
variables are bound and can be any string of letters. The {\tt
  let} constructor allows the use of macros in the writing of the
lexicon. The {\tt transformation} constructor implements unary rules
for type raising. Type raising rules are just allowed for some
categories and their application is controlled during execution.
The {\tt lex} constructor is used for each lexical entry.  In
this experiment we do not impose any type discipline (HPSG style) in the
feature structures themselves\footnote{Neither the distinction between
  ``syntactic'' and ``semantic'' features is made.}. If we assign to
each part of speech a feature structure, then an associated feature
description will be of type $fs \rw \ldots \rw {\bf bool}$.  For
instance, if we assign the type $fs\rw{\bf bool}$ to ``John'', with
semantics $\lambda s.s=john$, and assign the type $fs\rw fs \rw {\bf
  bool}$ to ``runs'', with semantics $\lambda x.\lambda s. s.reln=run
\wedge s.arg1=x$, the sentence ``John runs'' would have the type
$fs\rw{\bf bool}$ and semantics $\lambda s. s.reln=run\, \wedge\,
s.arg1=john$. Once more we note that the use of partial descriptions
allows us to express directly, the relations between the several
constituents. The semantic used is inspired in the ones
in \cite{PereiraS87}.
\begin{figure}
  \begin{center}
    \leavevmode
{\tt \small
\begin{tabbing}
Base\_\=Categories \qquad\qquad\=\% Define the set of base categories \\
\>      s = fs$\mathtt{->}$bool,       \>\% and their types \\
\>      iv = fs$\mathtt{->}$fs$\mathtt{->}$bool, \\
\>      np = s/iv, \\
\>      tv = iv/np, \\
\>      dv = tv/np, \\
\>      n = fs$\mathtt{->}$fs$\mathtt{->}$bool, \\
\>      det = np/n, \\
\>      pp= fs$\mathtt{->}$ bool;\\      
transformation          \>\>\% define a type raising rule\\
\>      np = (s/np)/(iv/np) : $\lapp$S $\lapp$Vt $\lapp$C. S (Vt C); \\
\%\%\%\%\%\%\%\% some useful abbreviations\\
\% agreement specifications\\
let 3RD\_SG =  $\lapp$X. X.pers=p3 \& X.nb=sg;\\
let NOT\_3RD\_SG =  $\lapp$X. X.pers$\lapp$=p3 $\mathtt{\mid}$ X.nb$\lapp$=sg;\\
let ANY = $\lapp$X. X=X;\\
\% proper nouns (generalized quantifier type)\\
let PN(W) = $\lapp$P.$\lapp$s. s.quant=exists\_one \& s.arg.reln=naming \\
\>   \&   s.arg.arg1=W \& 3RD\_SG(s.arg) \& P s.arg s.pred ;\\
\% common nouns (AGR is an agreement)\\
let CN(W,AGR) = $\lapp$s. s.reln=W \& s.arg1=x \& AGR s;\\
\%determiners\\
let DET(Q,AGR) = $\lapp$N. $\lapp$P. $\lapp$s. s.quant=Q \& AGR s.var \&\\
\>            N s.var s.range \& P s.var  s.scope;\\
\% intransitive verbs \\
let IV(W,AGR) = $\lapp$s.$\lapp$p. p.reln=W \& p.arg1=s \& AGR s;\\
\% transitive verbs (Obj is the semantics of the object)\\
let TV(W,AGR) = $\lapp$Obj. $\lapp$su.$\lapp$p. Obj ($\lapp$o
$\lapp$q. q.reln=W \& q.arg1=su \& q.arg2=o) p;\\
let V\_PP(W,AGR) = $\lapp$SS. $\lapp$su $\lapp$ s. SS s.arg2 \&
s.reln=W \& s.arg1=su \& AGR su; \\
\%ditransitive verbs\\
let DV(W,AGR) =  $\lapp$Ci.  $\lapp$Cs.  $\lapp$subj.
$\lapp$si. Cs ( $\lapp$ind.  $\lapp$s. Ci ( $\lapp$obj
$\lapp$p. p.reln=W \& \\
\> p.arg1=subj\& p.arg2=obj \& p.arg3=ind s) si \& AGR subj;\\

\%\%\%\%\%\%\%\%\%\%\%\%\%\% lexicon\\
lex a,    det,   DET(exists\_one,3RD\_SG); 
lex every,    det,   DET(all,ANY);\\
lex book, n,     CN(book,3RD\_SG); 
lex man, n,     CN(book,3RD\_SG);\\
lex john,  np, PN(john);
lex mary,  np, PN(mary);\\
lex died,  iv, IV(die,3RD\_SG); lex loves, tv, TV(love,3RD\_SG);\\
lex read, tv, TV(read,ANY);
lex said, iv/pp, V\_PP(say,ANY);\\
lex gave, dv, DV(give,ANY);\\
lex that, pp/s, $\lapp$s.s; \\
\% coordination
lex and, s$\lapp$(s/s), $\lapp$S1$\lapp$S2$\lapp$s.
s.type=coord \& S1 s.arg1 \& S2 s.arg2;\\ 
lex and, np$\lapp$((tv$\lapp$iv)/np), $\lapp$NP1$\lapp$NP2$\lapp$VT.
$\lapp$subj$\lapp$s. s.type=coord \&\\ 
\> VT NP1 subj s.arg1 \& VT NP2 subj s.arg2;\\
 lex and, iv$\lapp$(iv/iv), $\lapp$V1$\lapp$V2.
$\lapp$subj.$\lapp$s.\\ \> s.type=coord \& V1 subj s.arg1 \& V2 subj
s.arg2;\\ 
lex and, np$\lapp$(np/np),
$\lapp$NP1$\lapp$NP2$\lapp$VT$\lapp$s. s.type=coord \& \\
\> NP1 VT s.arg1 \& NP2 VT s.arg2;
\end{tabbing}
}
  \end{center}
  \caption{Sample grammar}
\label{fig1}
\end{figure}
\subsubsection{Processing}
\CCLG is implemented in Prolog augmented with the constraint solver
for feature descriptions\footnote{So it can be seen as an instance of
  \CLPLFD.}. In this section we briefly describe this implementation.
Although the feature descriptions used in the grammar are untyped, a
type inference algorithm is used to infer types for each expression.
Moreover, for each lexical entry the type of the feature description
is checked with that of the category and whenever possible the {\em
  normal form} of the feature description is computed.  The inference
rules are build-in in the grammar processor.  Currently, we use a
bottom-up chart parser that builds a context-free backbone. Each edge
is a (Prolog) term $arc(Begin,End,Cat,Sref)$ where $Cat$ is the
category spanning from $Begin$ to $End$ and $Sref$ is the information
to be used to extract the semantic representation, and that reflects
how this edge was formed: if it was a lexical entry $Sref$ is a
reference to it; if it results from a left (right) application rule,
it is a pair of references for its daughters; if it results from a
unary rule, it is a pair of references to the initial category and to
that rule. When the parse trees are successful built, the semantic
representation is extracted and the constraint solver applied. These
two components can be interleaved in order to prune, as soon as
possible, inconsistent edges.  As is apparent from the sample grammar
(figure \ref{fig1}.)  the semantic representations can become very
cumbersome to write and visualize.  So a graphical ``workbench'',
based on a Tcl/Tk interface to Yap Prolog, was provided to edit
grammars and lexicon, as well as to visualize the parse trees and
semantic representations (as matrix boxes).

\subsubsection{An Example}
\begin{figure}\noindent
\begin{minipage}[b]{.46\linewidth}
{\centerline{\epsfxsize=100pt\epsfbox{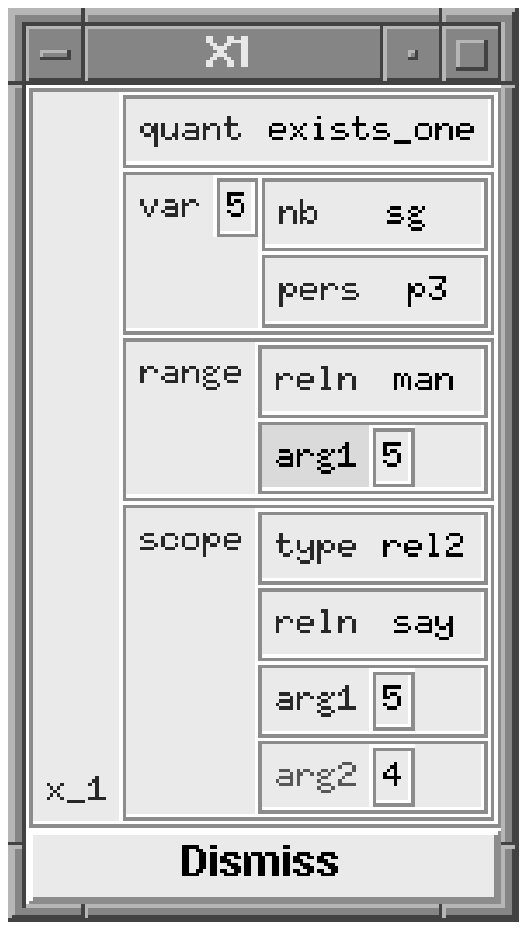}}}
\caption{Coordination inside relative} \label{w2}
\end{minipage}\hfill
\begin{minipage}[b]{.46\linewidth}
{\centerline{
\epsfxsize=125pt
\epsfbox{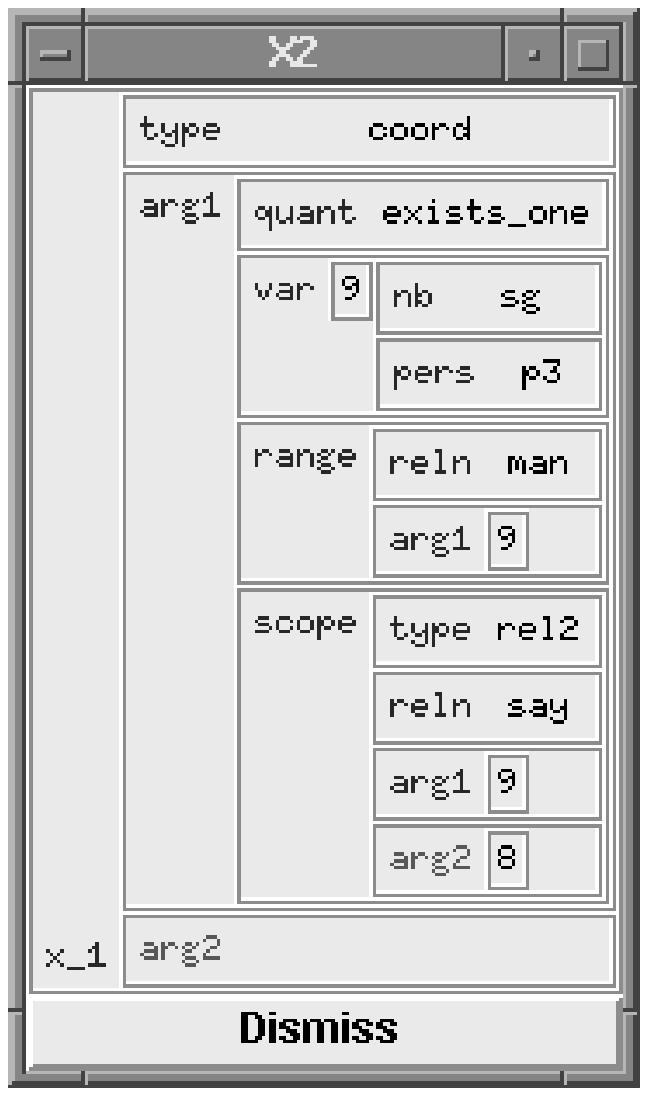}}}
\caption{Coordination wider scope }\label{w3}
\end{minipage}
\end{figure}

\begin{figure}
\noindent
\begin{minipage}[b]{.46\linewidth}
{\centerline{
\epsfxsize=125pt
\epsfbox{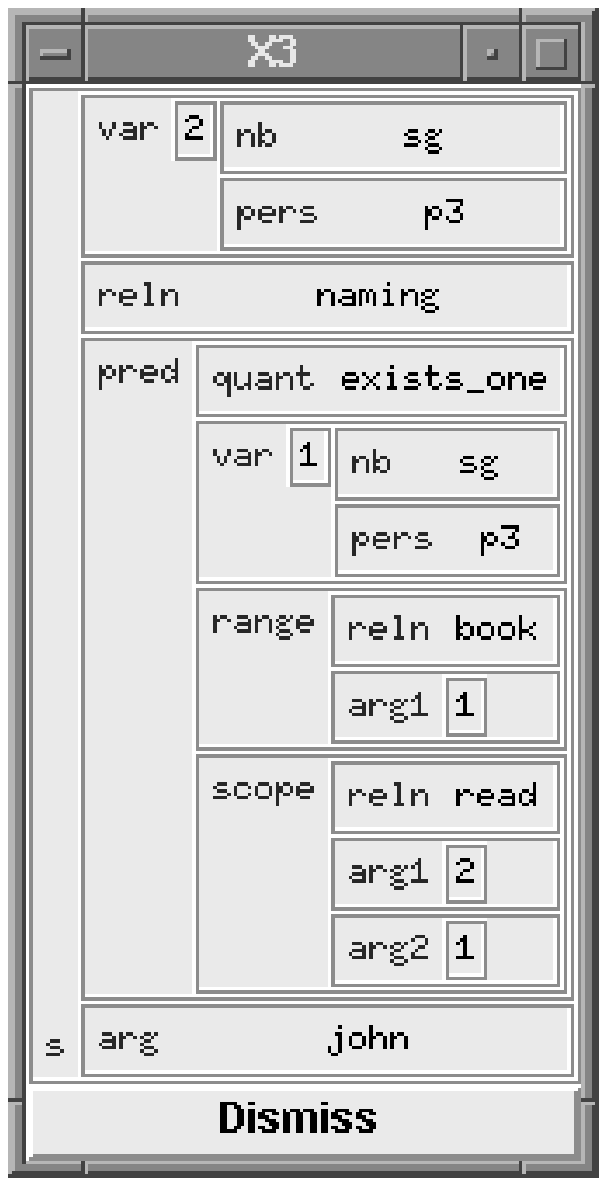}
}}
\caption{Semantics of ``john read a book''.}
\label{w4}
\end{minipage}\hfill
\begin{minipage}[b]{.46\linewidth}
{\centerline{\epsfxsize=100pt
\epsfbox{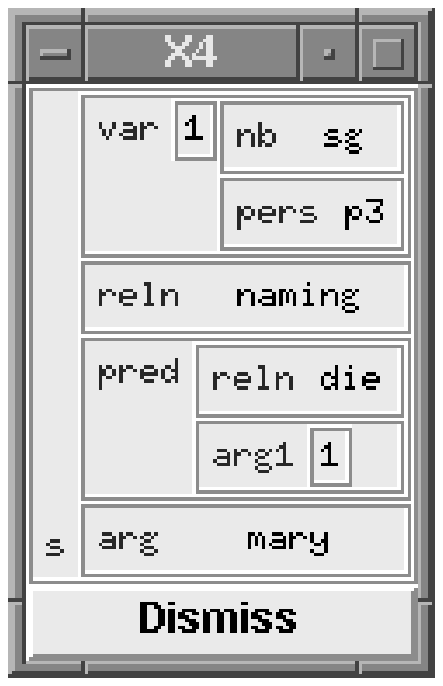}}}
\caption{Semantics of ``mary died''.}\label{w5}
\end{minipage}

\end{figure}

\begin{figure}
{
{\centerline{
\epsfxsize=300pt
\epsfbox{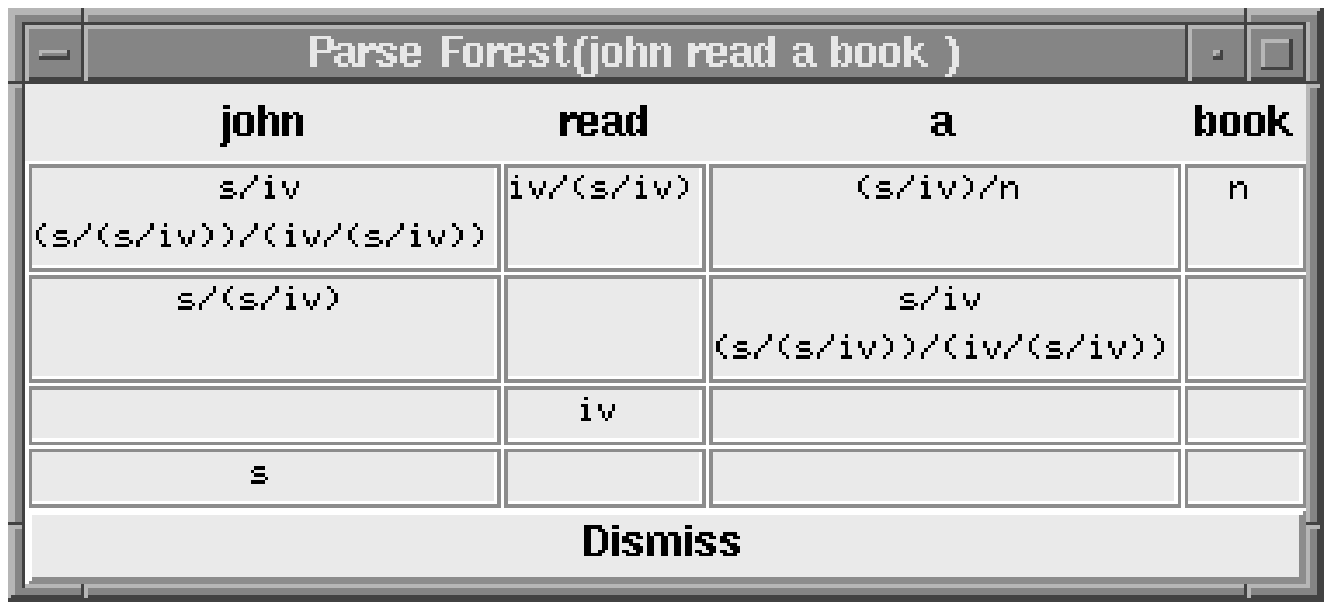}}}
}
\caption{A parse forest }
\label{w6}
\end{figure}

As an example we analyze the parsing of the sentence ``{\tt a man said
  that john read a book and mary died}''. There are two possible parse
trees of this sentence, one with the coordination in the scope of the
relative clause and other with a wider scope.  The semantic
representation of this sentence will be a feature description
$\mathtt{ \lapp x\_1.X1|X2}$ where {\tt X1} and {\tt X2} are partially
represented in figures \ref{w2}. and \ref{w3}.. Figures \ref{w4}.  and
\ref{w5}. show the semantics of the sentences ``{\tt john read a
  book}'' and ``{\tt mary died}'', respectively.  In the feature
description {\tt X2} (figure \ref{w3}.) the former semantics is
identified with the value of {\tt x\_1.arg1.scope.arg2} and the latter
is identified with the value of {\tt x\_1.arg2}. In the feature
description {\tt X1} (figure \ref{w2}.) the value of {\tt
  x\_1.scope.arg2} is the feature structure corresponding to the
coordination of the these two sentences.  As remarked in the previous
section, the parsing process first builds a parse forest using only
the categories of lexical items and the inference (and unary) rules
for syntactic categories. The parse tree of sentence we are
considering is too large to be considered here, so figure \ref{w6}.
shows only the parse forest of ``{\tt john read a book}''. In the
first row we have the syntactic categories of each lexical item (given
in the lexicon or derived by a unary rule).  In the following rows each
entry corresponds to the possible ways of deriving a category spanning
a portion of the input sentence. For instance, the category
$\mathtt{iv}$ can be derived in the third row from
$\mathtt{iv/(s/iv)}$ and $\mathtt{s/iv}$, spanning ``{\tt read a
  book}''.  Then for each parse tree that spans the whole sentence with root
category $s$, the semantic representation of the constituents are
combined and if the constraint solver does not produce $\bf false$, a
semantic representation is derived.

\noindent

\section{Final Remarks}\label{fr}
The current implementation of \CCLG shows the practical feasibility of
using higher order feature structure descriptions as semantic
representations.  This reflects the fact that the complexity of the
satisfiability problem for higher order feature descriptions is
essentially the same as for feature logics. We should also point out
that the good performance of the system results in part from its
hybrid nature where a categorial grammar with atomic base categories
is used to guide parsing. Some more  toy English
grammars where written that can handle some kinds of discontinuity, modifiers and
quantifier scope. However, the introduction of a type
discipline and more general treatment of recursive lexical rules
(\cite{BoumaN94}) must be considered, in future work. On the other
hand, most recent
developments of categorial grammars are based on the Lambek calculus
\cite{Lambek58,Moortgat88,Morril94} (an
intuitionist fragment of Linear Logic). Some implementations for the
propositional fragment are based on chart parsers
\cite{Konig94,Hepple92} and we conjecture that \LFD calculus can be
successfully used in such a systems, for process semantic
representations. 
From an implementational perspective it would be helpful  to study how current
techniques employed in functional programming implementations, namely the use of combinators,
can be imported for improve the computation of $\beta$-reductions. 
\subsubsection{Acknowledgments}
The authors would like to thank Sabine Broda and the anonymous
reviewers for their valuable comments on an earlier draft of this
paper.

\end{document}